\documentclass[10pt, conference, letterpaper]{ieeeconf}
\IEEEoverridecommandlockouts
\usepackage{multicol}
\usepackage[bookmarks=true]{hyperref}
\usepackage{cite}
\usepackage{graphics}
\usepackage{epsfig}
\usepackage{amsfonts}
\usepackage{amssymb,amsmath}
\usepackage{units}
\usepackage{bm}
\usepackage{multirow}
\usepackage{subfigure}
\usepackage{times}
\usepackage{rotating}
\renewcommand{\baselinestretch}{1}
\usepackage{algpseudocode}
\usepackage{blindtext}
\title{\LARGE \bf Novel Spring Mechanism Enables Iterative Energy Accumulation under Force and Deformation Constraints}
\author{Cole A. Dempsey and David J. Braun
\thanks{C. A. Dempsey and D. J. Braun are with the Department of Mechanical
Engineering, Vanderbilt University, Nashville, Tennessee 37235, USA.\newline
\indent This work was supported in part by a Seeding Success Grant provide by Vanderbilt University and a National Science Foundation CAREER Award (Grant No. 2144551). The authors gratefully acknowledge the support.\newline
\indent This paper has supplementary downloadable multimedia material. The video demonstrates the energy accumulation method presented in this work.\newline
\indent E-mail: {\tt cole.a.dempsey@vanderbilt.edu}\newline
\indent E-mail: {\tt david.braun@vanderbilt.edu}\newline
\indent © 20xx IEEE. Personal use of this material is permitted. Permission from IEEE must be obtained for all other uses, in any current or future media, including reprinting/republishing this material for advertising or promotional purposes, creating new collective works, for resale or redistribution to servers or lists, or reuse of any copyrighted component of this work in other works.
}}
\begin{document}
\maketitle
\begin{abstract}
Springs can provide force at zero net energy cost by recycling negative mechanical work to benefit motor-driven robots or spring-augmented humans. 
However, humans have limited force and range of motion, and motors have a limited ability to produce force.
These limits constrain how much energy a conventional spring can store and, consequently, how much assistance a spring can provide.
In this paper, we introduce an approach to accumulating negative work in assistive springs over several motion cycles. We show that, by utilizing a novel floating spring mechanism, the weight of a human or robot can be used to iteratively increase spring compression, irrespective of the potential energy stored by the spring. 
Decoupling the force required to compress a spring from the energy stored by a spring advances prior works, and could enable spring-driven robots and humans to perform physically demanding tasks without the use of large actuators. 
\end{abstract}

\section{Introduction}
Springs can enable robots actuated by motors \cite{Paluska2006,Braun2013,Plooij2016,Vu2016,Mazumdar2017,Kolvenbach2019} and humans ``actuated by muscles" \cite{Hollander2006,Wang2011,Collins2015,Braun2016,Braun2019,Sutrisno2020,Shin2021,Wang2021} to perform physically demanding tasks with reduced force requirements from the actuators.
In most applications, a spring is compressed slowly over a longer period of time, while the energy stored by the spring is released rapidly \cite{Hawkes2022}. 
In this way, the spring provides power amplification beyond what a motor driven robot or ``muscle actuated" human can do without the assistance of a spring. However, the energy stored by a spring is limited by the maximum force used to compress the spring.
Consequently, the maximal force that a robot or human can generate limits the amount of energy a spring can store, and the level of assistive benefit a spring can provide. This limitation may be alleviated by leveraging the energy storage ability of springs over multiple loading and unloading cycles instead of a single cycle.

In mechanical resonance, the benefit of springs is leveraged over multiple cycles of energy storage and release, instead of a single cycle  \cite{Okubo1996,Braun2011,Velasco2015,Haldane2016,Sutrisno2019,Mathews2022}. A familiar example is a pogo-stick, essentially a spring in series with the human legs, that allows the user to jump repeatedly to accumulate energy and reach jump heights much greater than in a single jump \cite{Hansburg1955}.
To accomplish such a feat, the pogo-stick relies on iteratively increasing the kinetic energy of the human through multiple jumps. This increase in kinetic energy is required to generate the large contact forces to compress the pogo-stick spring and thereby increase the energy stored by the spring. However, large forces are challenging for humans and robots to generate without increasing their kinetic energy.

In this paper, we present a method and a device for iteratively accumulating energy using only the static gravitational force provided by the mass of a spring-driven robot or the mass of a human augmented with a spring leg exoskeleton. The method utilizes the repeated application of a constant static force that is independent of the energy stored by the spring. The method also relies on a new device, which belongs to the class of floating spring mechanisms recently introduced in \cite{Kim2021}. The device is an energetically passive variable stiffness spring which automatically adjusts its stiffness to ensure that a constant force can compress the spring regardless of how much energy is stored by the spring.

The working principle of the device is demonstrated using a point mass model of the human squatting with a lower-limb spring leg exoskeleton in Section~\ref{section:2main}. The energy accumulation method is subsequently described in Section~\ref{section:mechtheory}. Then, the prototype of the device is detailed in Section~\ref{subsection:prototype}. Finally, the experimental validation of the proposed energy accumulation method is reported in Section~\ref{subsection:mechexperiment}.

\begin{figure*}[ht]
	\centering
	\includegraphics[width=1.0\textwidth]{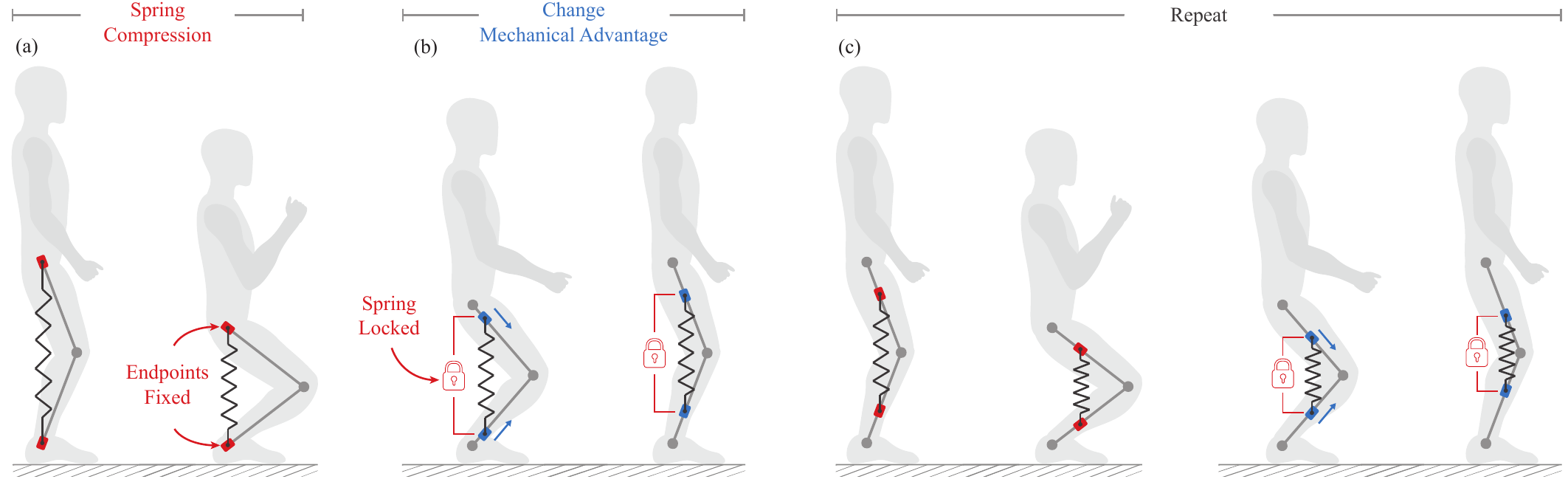}
	\caption{Repeated squatting with a floating variable stiffness spring. (a) The end points of the spring are fixed (red) while the user compresses the spring with a squat. (b) The end points of the spring are free (blue) while the spring is locked. The mechanical advantage of the human over the spring is increased as the human stands and the spring shifts towards the knee joint. (c) By repeating the energy accumulation cycle (a)-(b), the user can iteratively increase the energy stored by the spring.}
	\label{fig:Proc}
\end{figure*}

\section{Energy Accumulation using Springs}\label{section:2main}
Let us consider a simple energy accumulation task where the human is augmented with a spring exoskeleton attached parallel to the legs, see Fig.~\ref{fig:Proc}. In this task, the user compresses the spring by repeatedly squatting with the exoskeleton. The energy stored by the spring is retained by locking the spring at the bottom of each squat. As the human returns to the standing height, the spring shifts to a new configuration that grants the user a greater mechanical advantage over the spring for the next iteration. The greater mechanical advantage ensures the user can compress the spring at the beginning of each squat cycle until a desired amount of energy is accumulated in the spring. In the described iterative energy accumulation process, the force required by the human to achieve full spring compression is independent of the energy stored by the spring. 

In the remainder of this section, we further explore the example squatting task using a simple spring mass model of the human augmented with a conceptual lower-limb variable stiffness spring exoskeleton.

\subsection{Model} \label{subsection:fixed}
\begin{figure}[ht!]
	\centering
	\includegraphics[width=1.0\columnwidth]{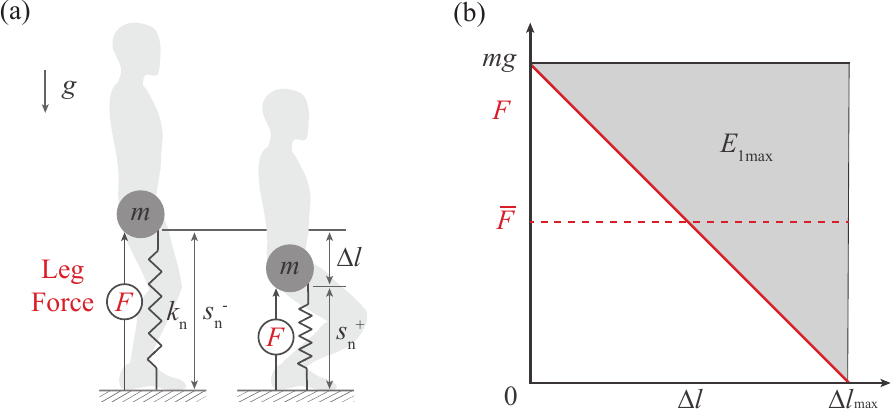}
	\caption{Model of the human augmented with a lower-limb exoskeleton. (a) Mass-spring system. The leg deformation is described by $\Delta l$. The body mass is supported by a spring with stiffness $k_n$ and deformed length $s_n^{\pm}$; the superscripts ${\pm}$ denote the pre-squat and post-squat spring lengths, respectively, and the subscript $n$ denotes the number of squats performed during repeated squatting. (b) An example force-deflection of the human leg $F$ (red) that leads to the average leg force $\bar{F}=\frac{1}{2}mg$ is shown with the red line. The maximum amount of energy accumulated during one squat $E_{1\max}$ is shown with the dark area.}
	\label{fig:mass_spring}
\end{figure}

The human and the spring-leg exoskeleton are abstracted into the model shown in Fig.~\ref{fig:mass_spring}a. We consider a single squat, starting from an upright standing position and ending at the fully squatted position. Due to the geometric constraint of the leg, the leg deformation during a squat is given by
\begin{align}
	&\Delta l \in[0,\Delta l_{\max}].
\end{align}

Because the human leg can push but cannot pull against the ground while squatting, the force exerted by the leg on the center of mass must be positive,
\begin{align}
	&F \geq 0.
\end{align}

Finally, we assume that the human legs can produce enough force to stand up after each squat without the support of the exoskeleton. Any leg force that can overcome the weight of the human $F\geq mg$ suffices this assumption.

An example of the human leg force that enables squatting from a standing position to the equilibrium position is shown in Fig.~\ref{fig:mass_spring}b.
At standing the human limbs fully support the mass such that $F=mg$ and the spring-leg exoskeleton does not provide any force. In the fully squatted position, the human limbs may support the mass with a force $F\in[0,mg)$ while the spring leg provides the rest of the force required to keep the center of mass in static equilibrium,
\begin{align}
	mg = F + k \Delta l_{\max}.
	\label{eqn:fb_fixed}
\end{align}

At the bottom of the squat, the energy stored by the spring leg depends on the human limb force. Assuming an average limb force $\bar{F}$, the energy stored by the spring is given by:
\begin{align}
    \frac{1}{2}k \Delta l_{\max}^2 = (mg - \bar{F}) \Delta l_{\max}.
    \label{eqn:E1_fixed}
\end{align}

In order to simultaneously satisfy (\ref{eqn:fb_fixed}) and (\ref{eqn:E1_fixed}), the average force of the human leg $\bar{F}$ during the squat must satisfy the following condition:
\begin{align}
\bar{F} = \frac{1}{2} (mg + F).
\label{eqn:Fbar_fixed}
\end{align}

According to (\ref{eqn:E1_fixed}) and (\ref{eqn:Fbar_fixed}), the maximum amount of energy that can be stored in the spring during a single squat is given by:
\begin{align}
	E_{1\max} = \frac{1}{2} mg \Delta l_{\max},
	\label{eqn:E1max}
\end{align}

Relation (\ref{eqn:E1max}) directly shows that the amount of energy accumulated in the spring is restricted by the range of motion and the limited gravitational force available to compress the spring. In the next section, we introduce a novel spring mechanism that can alleviate the aforementioned limitation.

\section{Cyclic Energy Accumulation by a Floating Spring Variable Stiffness Leg}\label{section:mechtheory}
There are three main practical challenges of realizing energy accumulation beyond a single squat: 

(i) The mechanism must provide increased mechanical advantage to compress the spring, such that the same force can be used to compress the spring even as it stores more energy. 

(ii) The mechanism must provide controllable coupling between the spring and the leg, such that the same leg deformation can be used to input energy into the spring in subsequent squats independent of how much energy is stored by the spring.

(iii) To ensure efficient energy accumulation, the two prior tasks should be accomplished while maintaining the energy stored in the spring between subsequent iterations. 

\begin{figure}[t]
	\centering
	\includegraphics[width=1.0\columnwidth]{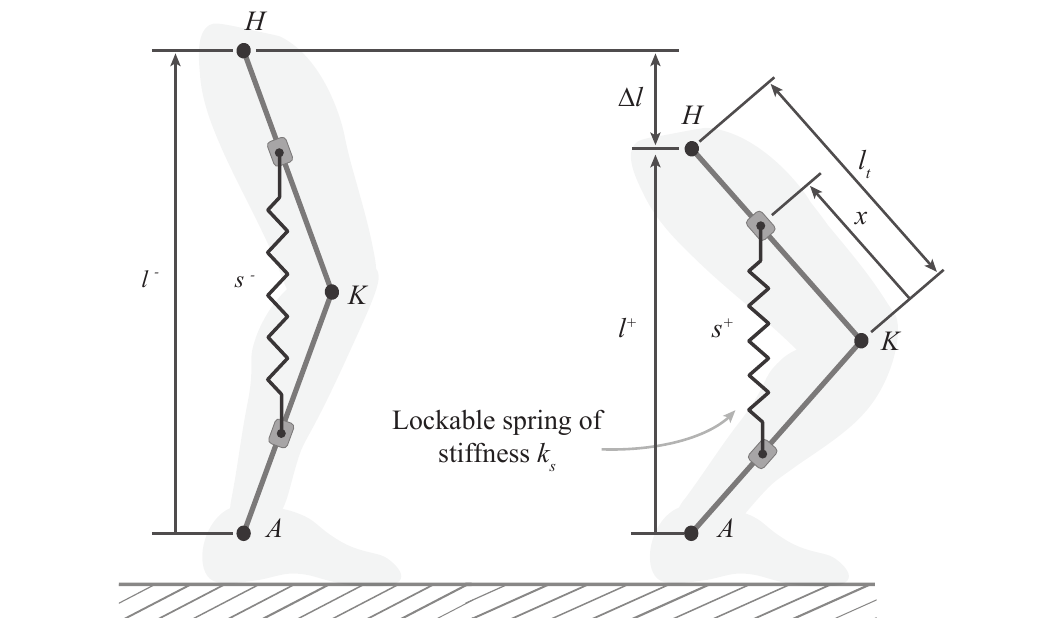}
	\caption{Model of the variable stiffness floating spring-leg. The leg segments $\overline{HK}$ and $\overline{KA}$ are equal in length. The spring is assumed to remain vertical ($x$ is constant) as the leg deforms by $\Delta l$.}
	\label{fig:Geo}
\end{figure}

In order to address the aforementioned challenges, a special case of the floating spring mechanism proposed in \cite{Kim2021} is presented (Fig.~\ref{fig:Geo}). In the original design, the mechanical advantage of the human over the spring was manipulated by changing the orientation of the spring. In the proposed design, the spring maintains a vertical orientation but shifts towards or away from the knee joint to change the mechanical advantage.
Due to this difference, the proposed design always maintains positive force-deflection behavior.
Both designs, however, alter mechanical advantage by controlling the endpoints of the spring while the spring is locked, which maintains the potential energy stored by the spring.
In turn, this ability to control the endpoints of the spring allows the leg deformation to be decoupled from the spring deformation, independent of the energy stored by the spring. 
Consequently, the proposed design addresses all three of the practical challenges (i)-(iii) mentioned above.

In what follows, we present the geometry that grants the spring leg mechanism energetically efficient variable stiffness behavior (Section~\ref{subsection:mech}), and describe the working principle of the mechanism (Section~\ref{subsection:both}) consistent with the three main requirements outlined above.

\subsection{Model} \label{subsection:mech}
Figure~\ref{fig:Geo} shows the floating spring variable stiffness leg for a single squat iteration. In the leg, points $H$, $K$ and $A$ coincide with the user's hip, knee, and ankle, respectively. The thigh and shank segments $\overline{HK}$ and $\overline{KA}$ are assumed to be of equal length $l_t$. The spring is also assumed to maintain its vertical orientation independent of the leg deformation. 
In the mechanism, the length of the spring $s$ is defined by leg length $l$ and the position of the spring $x$, 
\begin{align}
    &s = \frac{x}{l_t} l,
    \label{ls_ratio}
\end{align}
while the force required at the hip $F_{l}$ to compress the spring is defined by, 
\begin{align}
    F_{l} = \bigg(\frac{x}{l_t} \bigg) F_{s} = \bigg(\frac{x}{l_t} \bigg) k_s (s_0-s),
    \label{kl_stiff}
\end{align}
where $s_0$ is the uncompressed length of the spring and $k_s$ is the stiffness of the spring.

These relations suggest that by moving the spring towards the knee joint -- decreasing $x$ -- a small constant force $F_l$ could be used at the hip to compress the spring despite a potentially large spring force $F_s$, and consequently, the large amount of energy stored by the spring. 

The next section explores how the mechanism shown in Fig.~\ref{fig:Geo} could accumulate a large amount of energy when compressed by the weight of the human over multiple squats. 

\subsection{Cyclic Energy Accumulation} \label{subsection:both}
In order to predict the behavior of the mechanism beyond a single squat, we again consider the simple example of a human performing a repetitive squat task to accumulate energy in the spring as introduced in Section~\ref{section:2main}. 

First, similar to (\ref{eqn:fb_fixed}), we assume that the spring-leg supports the weight of the user at the end of each squat,
\begin{align}
    &mg = k_s\left ( \frac{x_n}{l_t} \right )\left ( s_0 - s_{n}^{+} \right ).
    \label{eqn:fb_mech}
\end{align}

To define the spring length at the end of the squat $s_n^+$, the spring location $x_n$ must be related to the spring length at the beginning of the squat $s_n^-$. The simple relation below follows from locking the spring length between the end of the previous squat and the beginning of the next squat (Fig.~\ref{fig:Proc}),
\begin{align} 
	s_{n-1}^{+} = s_{n}^{-}.
	\label{eqn:ss}
\end{align}
According to (\ref{eqn:ss}), the energy stored by the spring will be retained between squats, 

\begin{align} 
    \frac{1}{2}k_s\left ( s_{n-1}^{+} - s_0 \right )^2 = \frac{1}{2}k_s\left ( s_{n}^{-} - s_0 \right )^2.
	\label{eqn:E_mech}
\end{align}

Finally, using (\ref{ls_ratio}) and (\ref{eqn:ss}), we define a recurrence relation that predicts the position of the spring across squat iterations:
\begin{align}
    x_n = \frac{s_{n}^{-}}{s_{n-1}^{-}} x_{n-1}.
    \label{eqn:x_eqn}
\end{align}

Substituting (\ref{eqn:fb_mech}), (\ref{eqn:ss}), and (\ref{eqn:x_eqn}) into (\ref{kl_stiff}), we find that the force required to compress the spring at the beginning of the next squat is always lower than the constant gravitational force available to compress the spring, 
\begin{align}
    F_{ln}^{-} = \left ( \frac{s_{n}^{-}}{s_{n-1}^{-}} \right ) mg \; \leq \; mg.
    \label{eqn:Fln}
\end{align}

\begin{figure}[t!]
	\centering
	\includegraphics[width=\columnwidth]{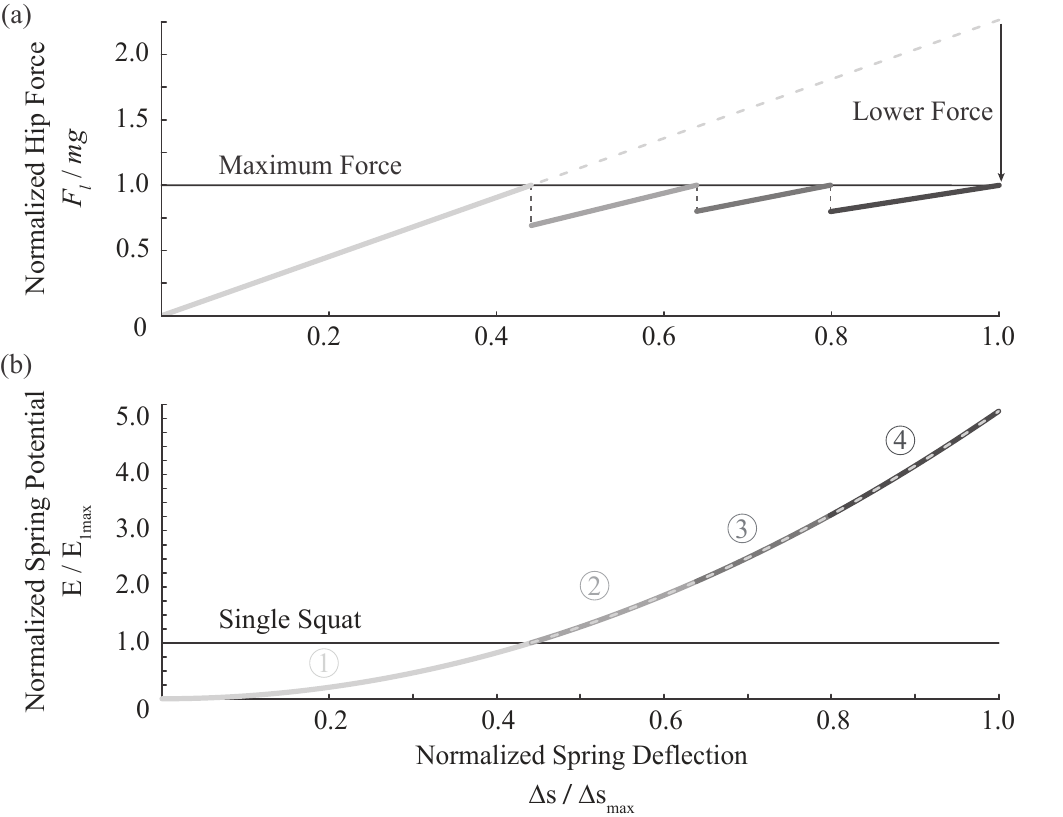}
	\caption{Simulated behavior for the spring-leg. (a) Force-deflection achieved by repeated squats, subject to a limited force (horizontal line), compared to the force required to achieve the same deformation in a single compression cycle (dashed line). The forces are normalized by the weight of the user (Section~\ref{subsection:fixed}). (b) Potential energy stored by the spring, normalized by the max amount of energy that can be accumulated in a single squat $E_{1\max}$ (\ref{eqn:E1max}) defined in (Section~\ref{subsection:fixed}). The four squats shown in the figure were the minimum number of squats necessary to achieve full spring deformation.}
	\label{fig:fd_mechPlot}
\end{figure}

Figure~\ref{fig:fd_mechPlot}a shows the force-deflection predicted during multiple squats, during which the maximal force provided by the human is bounded by the weight of the user (gray). This force is compared to the force required to achieve the same spring deformation in a single squat (gray dashed). Further, the vertical dashed lines show that the spring deformation is maintained between iterations, as required by (\ref{eqn:ss}). 

Figure~\ref{fig:fd_mechPlot}b shows the energy accumulation process for the iterative method (gray), with energetic potential normalized by the maximum energy that can be achieved in a single squat provided by the weight of the user, $E_{1\max}$, as defined in (\ref{eqn:E1max}).

We observe that the spring accumulates the same amount of energy through four squats (gray) than in a single squat (gray dashed) but with less than half of the required force (Fig.~\ref{fig:fd_mechPlot}a-b). 
Furthermore, we observe that the reduction in force translates to over five times more stored energy as compared to what can be accumulated in a single squat subject to the same maximum force (Fig.~\ref{fig:fd_mechPlot}b). 
Following the repeated squats, the spring can be reset to the initial mechanical advantage, $x=l_t$, where the accumulated energy can be released to provide double the assistive force as compared to the maximal force used to compress the spring (Fig.~\ref{fig:fd_mechPlot}a) and the energy stored by the spring when compressed with the maximal force in a single squat (Fig.~\ref{fig:fd_mechPlot}b).

\section{Prototype} \label{subsection:prototype}
In this section, a prototype of the floating spring mechanism introduced in Section~\ref{section:mechtheory} is presented.

\subsection{Device} \label{subsection:prot}
\begin{figure}[t!]
	\centering
	\includegraphics[width=\columnwidth]{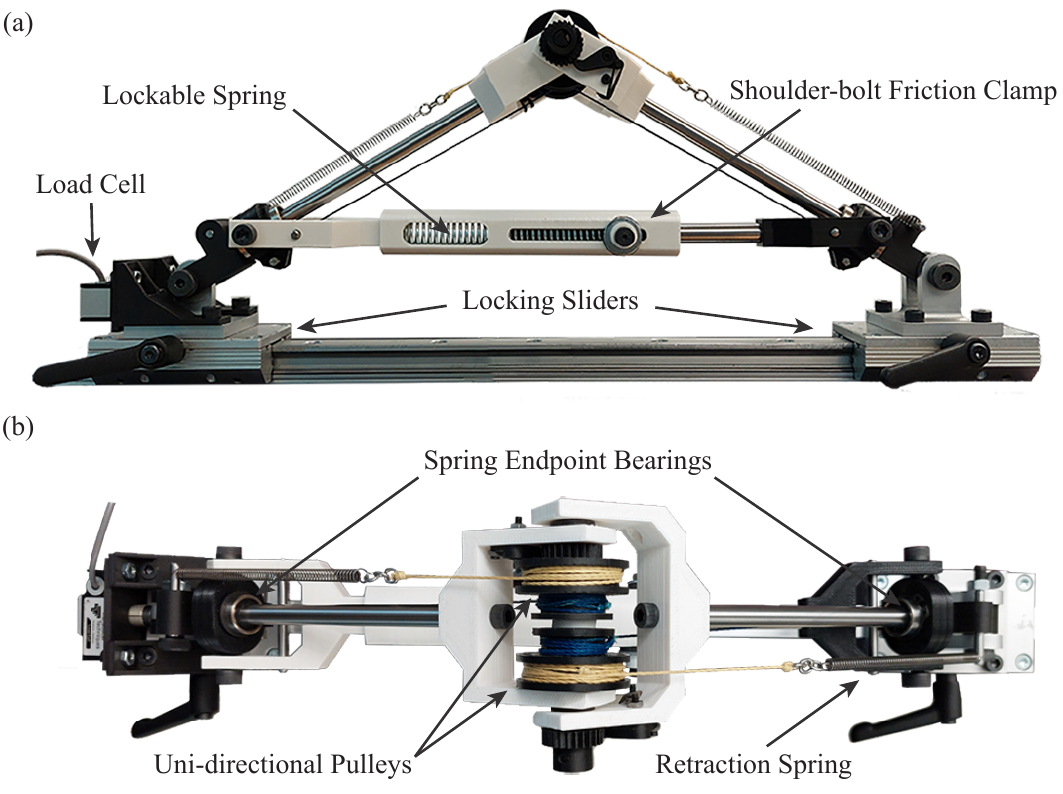}
	\caption{Prototype floating spring-leg mechanism. (a) Front view. (b) Top view. The leg segments are of relatively equal length of $205$~mm. The spring has a free length of approximately $114$~mm and a force-deflection rate of $0.9$~N/mm.}
	\label{fig:label_prot}
\end{figure}

Figure \ref{fig:label_prot} depicts the spring-leg prototype designed for the proposed energy accumulation task. The device consists of three major sub-assemblies: the leg structure, the compression spring, and the spring retraction mechanism. These sub-assemblies are detailed below.

First, the leg structure is formed by two linear shafts connected by brackets at the knee and pinned to create a hinge joint. The other ends of each shaft form the hip and ankle of the mechanism, as in Fig.~\ref{fig:Geo}.

Second, a compression spring is housed in a piston-cylinder assembly where each endpoint of the assembly connects to a linear ball bearing that slides freely along the leg shafts. To lock the spring axially, a shoulder bolt passes orthogonally through a hole in the piston and rides in a slot in the cylinder, as shown in Fig.~\ref{fig:label_prot}a. The cylinder features flat sides as an interface for the shoulder bolt, allowing continuous locking of the spring.

Finally, two uni-directional pulleys, each with two drums, are mounted in coincidence with the knee joint pin, as shown in Fig.~\ref{fig:label_prot}b. One pulley is keyed to the knee pin, while the other pulley spins freely on the knee pin. Both, however, can spin freely with respect to the knee brackets. The two drums on each pulley feature separate cables wrapped in opposing directions. One cable connects to an extension spring that provides a torque on the pulley, while the other cable is connected to an endpoint of the spring assembly. 
The system of pulleys serves to automatically shift the position of the spring assembly, and consequently change the mechanical advantage between the spring and the leg, between each compression cycle.

\subsection{Working principle} \label{subsection:stiffmod}
\begin{figure}[t!]
	\centering
	\includegraphics[width=\columnwidth]{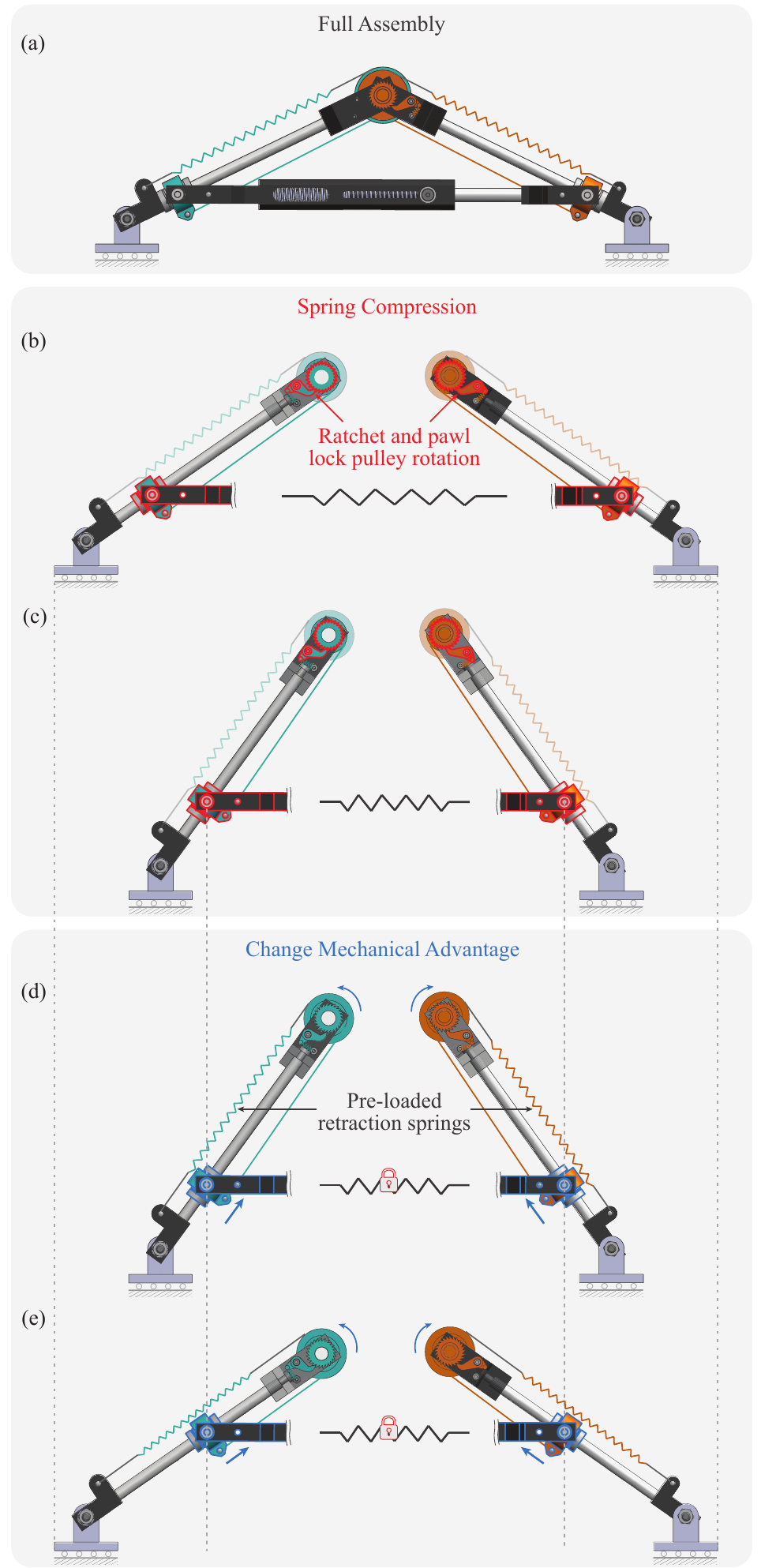}
	\caption{Compressing the spring and changing the mechanical advantage of the leg over the spring with the variable stiffness floating spring mechanism. (a) Front view of the CAD model. (b)-(c) Compressing the spring. During compression, the endpoints of the spring are fixed along the leg segments by locking the rotation of each respective pulley via a ratchet and pawl. (d)-(e) Changing the mechanical advantage of the leg over the spring. To change the mechanical advantage, the mechanism extends back to its initial configuration with the spring locked, allowing the pre-loaded retraction springs to rotate each pulley and simultaneously retract the spring endpoints towards the knee.}
	\label{fig:SLCAD_proc}
\end{figure}

In this section we examine the working principle of the prototype shown in Fig.~\ref{fig:label_prot}. 
Figure~\ref{fig:SLCAD_proc} shows the CAD model of the prototype, which features blue and orange colored parts to help distinguish which components are responsible for each endpoint of the spring assembly. As shown, each leg segment utilizes a pulley and dual cable assembly to move the endpoints of the spring assembly in unison.

During compression, see Fig.~\ref{fig:SLCAD_proc}b-c, the spring is unlocked and applies force on the bearing-mounted endpoints of the spring. Subsequently, the ratchet and pawl lock the rotation of each pulley with respect to their associated knee bracket. This cable-pulley setup then locks the position of the endpoints against the force of the spring.

Following the spring compression, the spring is locked, and the pre-loaded retraction springs apply a torque on their respective pulleys, see Fig.~\ref{fig:SLCAD_proc}d-e. Since the ratchet and pawl ensure the pulleys can only rotate in one direction, the torque on the pulley from the retraction spring tends to pull the endpoint of the spring assembly towards the knee joint. Therefore, as the mechanism returns to an initial configuration, the tension in the cables automatically shifts the endpoints of the spring towards the knee for a change in mechanical advantage before the next iteration. 

\begin{figure*}[t!]
	\centering
	\includegraphics[width=1.0\textwidth]{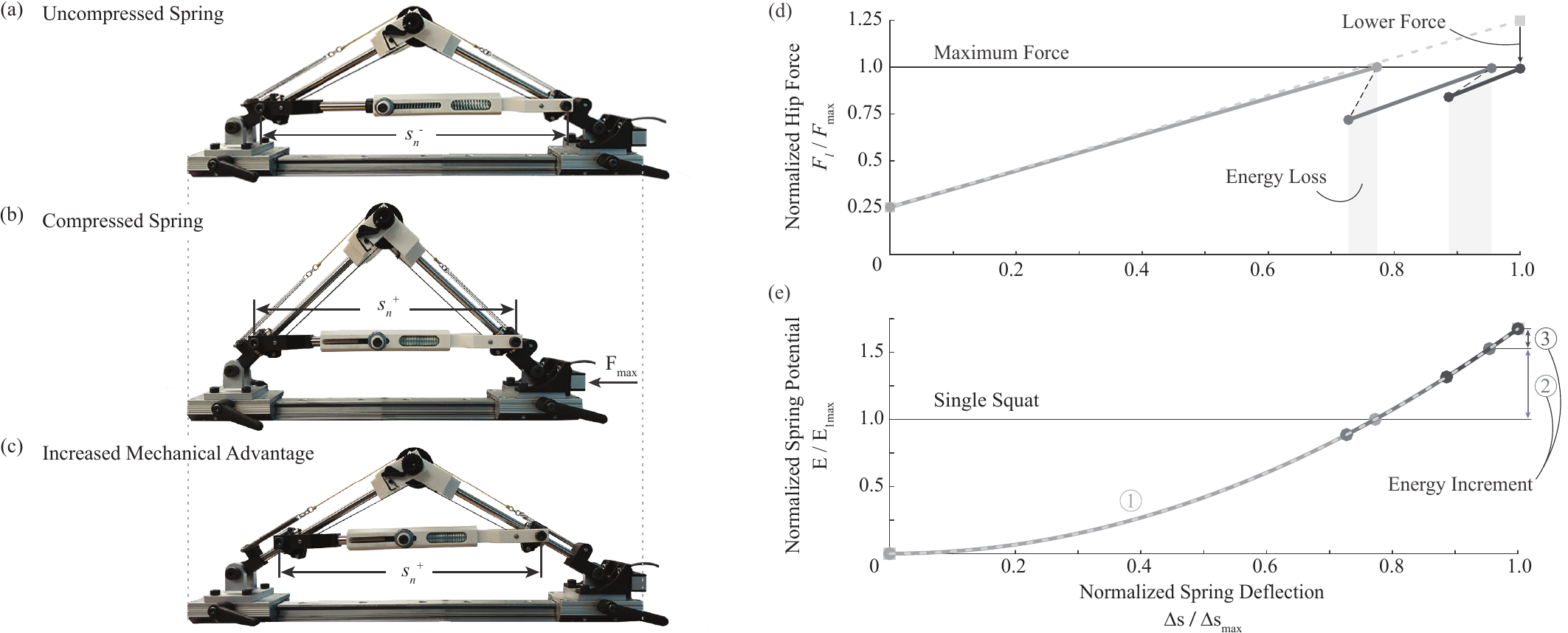}
	\caption{Experiment. (a)-(b) Example of one iteration performed during the experiment (solid lines in (d)). (c) Example of stiffness change between iterations (dashed lines in (d)). (d) Experimental force-deflection data for iterative energy accumulation using a limited force (solid lines). The data is normalized by the maximum force used to compress the spring. The maximum force  is compared to the force required to achieve the same spring deflection after one compression (dashed line). (e) Experimental energy accumulation data, normalized by the maximum energy that can be accumulated in a single squat subject to maximum force (solid lines). The energy stored by the spring after three iterations with the limited force is compared to the energy that could be stored after one compression using a larger force (dashed line).}
	\label{fig:mechExp}
\end{figure*}

\section{Experimental Validation} \label{subsection:mechexperiment}
In this section, we use the prototype presented in Section~\ref{subsection:prototype} to evaluate the theoretical predictions shown in Fig.~\ref{fig:fd_mechPlot}.

\subsection{Experimental Setup} \label{subsection:exp_setup}
The prototype, shown in Fig.~\ref{fig:label_prot}, was mounted on a mechanical breadboard via a linear rail. The ends of each leg segment were connected to lockable carriages. One carriage was locked in place, acting as the ankle joint fixing the foot of the mechanism to the ground, while the other carriage could move freely along the rail like the hip joint. A load cell (MLP-50, Transducer Techniques) was mounted to a flat plate on the free moving carriage to measure the force at the hip joint.

\subsection{Experimental Procedure} \label{subsection:exp_result}
To test the iterative energy accumulation process outlined in Section \ref{subsection:both}, the prototype was subject to the following experimental procedure.

First, with the spring unlocked, a force was applied to the load cell with constant velocity on the free slider until a pre-defined maximal force was reached. At that point, force was measured by the load cell.
After collecting the force data, the spring was locked axially by tightening the friction clamp and the length of the spring was measured. 
Next, the slider was unlocked and moved back to its initial position. As described in Fig.~\ref{fig:SLCAD_proc}d-e, moving the slider shifted the spring to a new configuration that granted greater mechanical advantage over the spring. The process described here was then repeated until maximum spring deformation was achieved.

\subsection{Experimental Result}
Figure~\ref{fig:mechExp} displays the experimental result. 

Figure~\ref{fig:mechExp}a, shows the force-deflection trend predicted in Section~\ref{subsection:both}. Force is observed to increase up to the maximum force, then decrease to allow another squat despite the increased potential of the spring. The decrease in force required to enable a new squat is accomplished by the ratchet, pawl, and pulley assembly, described in Section~\ref{subsection:stiffmod}. The iterative force-deflection behavior (solid lines) is compared to the force required to reach the same spring deflection in a single squat (dashed line).

Figure~\ref{fig:mechExp}b shows the iterative increase in spring potential. The results show that the spring accumulates the same energetic potential in three squats (solid lines) as compared to that accumulated in one squat (dashed line). However, similar to what was predicted in Section~\ref{subsection:both}, the mechanism reduces the amount of force necessary to accumulate the same amount of potential energy. In particular, the $25\%$ decrease in force observed in Fig.~\ref{fig:mechExp}a resulted in $75\%$ percent more energy accumulated by the spring compared to the energy one could store after a single squat when using the same maximal force. This result follows the same trend observed in Fig.~\ref{fig:fd_mechPlot}. 
Further, when the device is reset to the initial mechanical advantage $x=l_t$, it yields nearly 25\% more assistive force (Fig.~\ref{fig:mechExp}a) and 75\% more energy compared to the maximum force used to repetitively compress the spring and the associated energy stored by the spring when compressed by the maximum force in a single squat (Fig.~\ref{fig:mechExp}b).

While the experimental results were similar to the theoretical predictions, there were notable differences. For example, the mechanism exhibited roughly $84$ percent efficiency, due to the energy loss observed during the experiment. This energy loss is shown in Fig.~\ref{fig:mechExp}a by the black dashed lines not being vertical between iterations. The loss of energy can also be observed in Fig.~\ref{fig:mechExp}b, where the energy accumulated in the spring first decreased at the beginning of each new energy accumulation cycle.

Two main factors contributed to the observed energy loss; first, the cables used to couple the retraction springs to the spring assembly were not completely inextensible, and second, the ratchet and pawl only provide discrete locking positions of the spring endpoints along the leg shafts, and therefore introduced some amount of backlash.

One can also observe in Fig.~\ref{fig:mechExp}a that force is not initially zero, despite the zero initial spring potential, see Fig.~\ref{fig:mechExp}b. This initial force was due to pre-loading of the compression spring to mitigate slack in the pulleys and cables. Also, the forces created by the retraction springs tend to pull the spring endpoints towards the knee, which in turn creates a moment about the knee that wants to straighten the leg.

\section{Conclusion}
In this paper, we studied a model of a human performing a repetitive squat-to-stand task to accumulate energy in a lower-limb spring-leg exoskeleton. This task was used as a representative example of iterative energy accumulation in a spring under force and deformation constraints. We proposed a variable stiffness floating spring leg mechanism to demonstrate the novel force-deflection and energy storage behavior conjectured in this study. Our theoretical predictions were experimentally validated using a prototype variable stiffness floating spring mechanism.

The prototype supplements the floating spring technology introduced in one of our prior works \cite{Kim2021} by presenting a novel method for automatically adjusting the mechanical advantage of the human or a robot over a spring between compression iterations. The mechanism demonstrated the key novelty proposed in this work: a static gravitational force, provided by the mass of a human or robot, can be used to accumulate energy in a spring independent of the desired amount of energy stored by the spring.

The new capability of iterative energy accumulation using a limited static force, enabled by the method and device presented in this paper, could allow humans and robots with limited force capability and limited range of motion to perform physically demanding tasks, for example, to jump higher \cite{Sutrisno2019,Arm2019,Hawkes2022}, move faster \cite{Zhang2022}, or lift heavier objects \cite{Lamers2018}, by harnessing the energy stored in assistive springs. The results presented in this work pave the way towards novel designs of robot exoskeletons and spring-driven robots with enhanced energy storage capabilities.

\bibliographystyle{ieeetr}
\bibliography{bibliography}{}  
\end{document}